%% file: egpaper_for_review_iccv.tex
\documentclass[10pt,twocolumn,letterpaper]{article}
\usepackage{wacv}              
\usepackage[accsupp]{axessibility}
\usepackage{amsmath}
\usepackage{amssymb}
\usepackage{times}
\usepackage{epsfig}
\usepackage{amsfonts}
\usepackage{booktabs}
\usepackage{siunitx}
\usepackage{adjustbox}
\usepackage{tikz}
\usepackage{bm}
\usepackage{adjustbox}
\usepackage{dashrule}
\usepackage{mathtools}
\usepackage{makecell}
\usepackage{colortbl}
\usepackage{times}
\usepackage{graphicx}
\usepackage{algpseudocode}

\usepackage{cite} 
\usepackage{float}
\usepackage{enumitem}
\usepackage{dsfont}
\usepackage{array}
\usepackage{comment}
\usepackage{color}
\usepackage{wrapfig, lipsum}
\usepackage{capt-of}
\usetikzlibrary{babel}
\usepackage{tikz}
\usepackage{ctable}
\usepackage{graphicx}
\usepackage{amsmath}
\usepackage{comment}

\include{rvc-notation}

\usepackage[
  separate-uncertainty = true,
  multi-part-units = repeat
]{siunitx}

\definecolor{blueblack}{RGB}{0, 108, 173}
\definecolor{taborange}{RGB}{235, 127, 14}
\definecolor{tabgreen}{RGB}{30, 160, 30}
\definecolor{tabpurple}{RGB}{128, 103, 189}

\definecolor{sol_light_blue}{RGB}{38, 139, 210}
\definecolor{sol_blue}{RGB}{38, 139, 210}
\definecolor{nord_blue}{RGB}{38, 139, 210}
\definecolor{sol_green}{RGB}{163, 190, 140}
\definecolor{sol_red}{RGB}{220, 50, 47}
\definecolor{nord_red}{RGB}{250, 190, 192}
\definecolor{nord_green}{RGB}{163, 190, 140}

\definecolor{nordblack}{RGB}{46, 52, 64}
\definecolor{nordred}{RGB}{191, 97, 106}
\definecolor{magenta}{RGB}{215, 10, 185}
\definecolor{nordgreen}{RGB}{163, 190, 140}
\definecolor{nordblue}{RGB}{94, 129, 172}
\definecolor{nordpurple}{RGB}{180, 142, 160}

\DeclareMathOperator*{\argmin}{arg\,min}

\usepackage[pagebackref,breaklinks,colorlinks]{hyperref}

\usepackage[capitalize]{cleveref}
\crefname{section}{Sec.}{Secs.}
\Crefname{section}{Section}{Sections}
\Crefname{table}{Table}{Tables}
\crefname{table}{Tab.}{Tabs.}

\def\wacvPaperID{214} 
\def\confName{WACV}
\def\confYear{2024}

\begin{document}

\title{OptFlow: Fast Optimization-based Scene Flow Estimation without Supervision}

\author{Rahul Ahuja \quad Chris Baker \quad Wilko Schwarting\\
ISEE AI\\
{\tt\small \{rahulahuja, chrisbaker, wilko\}@isee.ai}}
\maketitle

\begin{abstract}
   Scene flow estimation is a crucial component in the development of autonomous driving and 3D robotics, providing valuable information for environment perception and navigation. Despite the advantages of learning-based scene flow estimation techniques, their domain specificity and limited generalizability across varied scenarios pose challenges. In contrast, non-learning optimization-based methods, incorporating robust priors or regularization, offer competitive scene flow estimation performance, require no training, and show extensive applicability across datasets, but suffer from lengthy inference times.

In this paper, we present OptFlow, a fast optimization-based scene flow estimation method. Without relying on learning or any labeled datasets, OptFlow achieves state-of-the-art performance for scene flow estimation on popular autonomous driving benchmarks. It integrates a local correlation weight matrix for correspondence matching, an adaptive correspondence threshold limit for nearest-neighbor search, and graph prior rigidity constraints, resulting in expedited convergence and improved point correspondence identification. Moreover, we demonstrate how integrating a point cloud registration function within our objective function bolsters accuracy and differentiates between static and dynamic points without relying on external odometry data. Consequently, OptFlow outperforms the baseline graph-prior method by approximately 20\% and the Neural Scene Flow Prior method by 5\%-7\% in accuracy, all while offering the fastest inference time among all non-learning scene flow estimation methods.
\end{abstract}
\section{Introduction}
\label{sec:intro}

Estimating 3D motion fields from dynamic scenes is a fundamental problem in computer vision with wide-ranging applications, from autonomous driving to scene parsing and object tracking. Scene flow estimation plays a vital role in many of these applications, enabling machines to perceive and navigate through their environments. For example, in autonomous driving, scene flow estimation helps vehicles understand the 3D structure and motion of the surrounding environment, which is essential for making safe and informed decisions. Similarly, in robotics, scene flow estimation assists robots in navigating through complex environments by providing a 3D understanding of the scene.

\begin{figure}[t!]
\centering
\includegraphics[width=\linewidth]{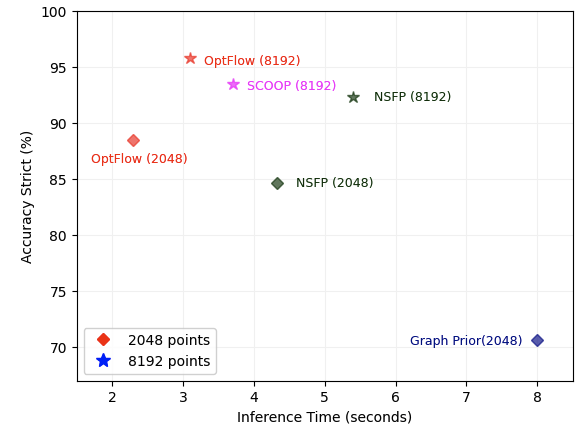}
\vspace{-10pt}
\caption{ \textbf{Graph depicting flow accuracy $Acc_5$  \vs inference time in seconds with 2048 and 8192 points used}. \textcolor{orange}{OptFlow} is the fastest algorithm while achieving state-of-the-art results among all the non-learning-based methods. The experiments were run on an NVIDIA Tesla T4 GPU.}
\label{fig:speed_vs_acc}
\end{figure}

Traditionally, state-of-the-art methods for scene flow estimation have relied heavily on image-based training methods using semantic depth. However, in recent years, interest in using lidar-based methods has grown. Learning-based methods~\cite{flownet3d, flownet3d++} pioneered this area, but these methods require large annotated datasets for training, which can be difficult and expensive to obtain in real-life scenarios. Even though many driving datasets~\cite{kitti, nuscenes, argoverse} are available, obtaining annotations for ground truth flow vectors can be a challenge.

This challenge has led to the development of large synthetic datasets, such as FlyingThings3D~\cite{ft3d}, which have emerged as an alternative means of training or pre-training models. However, a significant domain gap often exists between synthetic and real-world datasets. This gap has given rise to self-supervised scene flow estimation models~\cite{Mittal_2020_CVPR, pointpwc}, which reduce the domain gap by training the models on non-annotated datasets. However, these models often suffer from slow convergence and require a considerable amount of training time.

Optimization-based methods for scene flow estimation~\cite{graphprior, nsfp} optimize flow for each point cloud pair without using training data. However, their high accuracy is often accompanied by long processing times, which limits their practicality in some applications.

To address this issue, we present \textbf{OptFlow}, a fast non-learning optimization-based method. We introduce novel concepts such as the local correlation weight matrix, integrated ego-motion compensation, and adaptive max correspondence threshold limit, which significantly improves the convergence speed of our optimization method. Our method improves accuracy on real-world autonomous driving benchmark datasets by at least 20\% over the baseline method \cite{graphprior} and is competitive with the current state-of-the-art methods. 

Our work makes the following contributions:
\begin{enumerate}
    \item Point correspondence matching by incorporating a local correlation weight matrix for the target point cloud in the objective function. This helps with better aligning associated points and producing more accurate results. 
    \item An adaptive maximum correspondence threshold, which reduces noisy correspondences and further improves the quality of the estimates.
    \item An intrinsic point cloud matching transformation function based on ICP. This improves our flow estimates, increases the convergence speed, and helps distinguish static points from dynamic points.
    
\end{enumerate}
\begin{figure*}[t!]
\centering
\includegraphics[width=\textwidth, height=0.35\textheight]{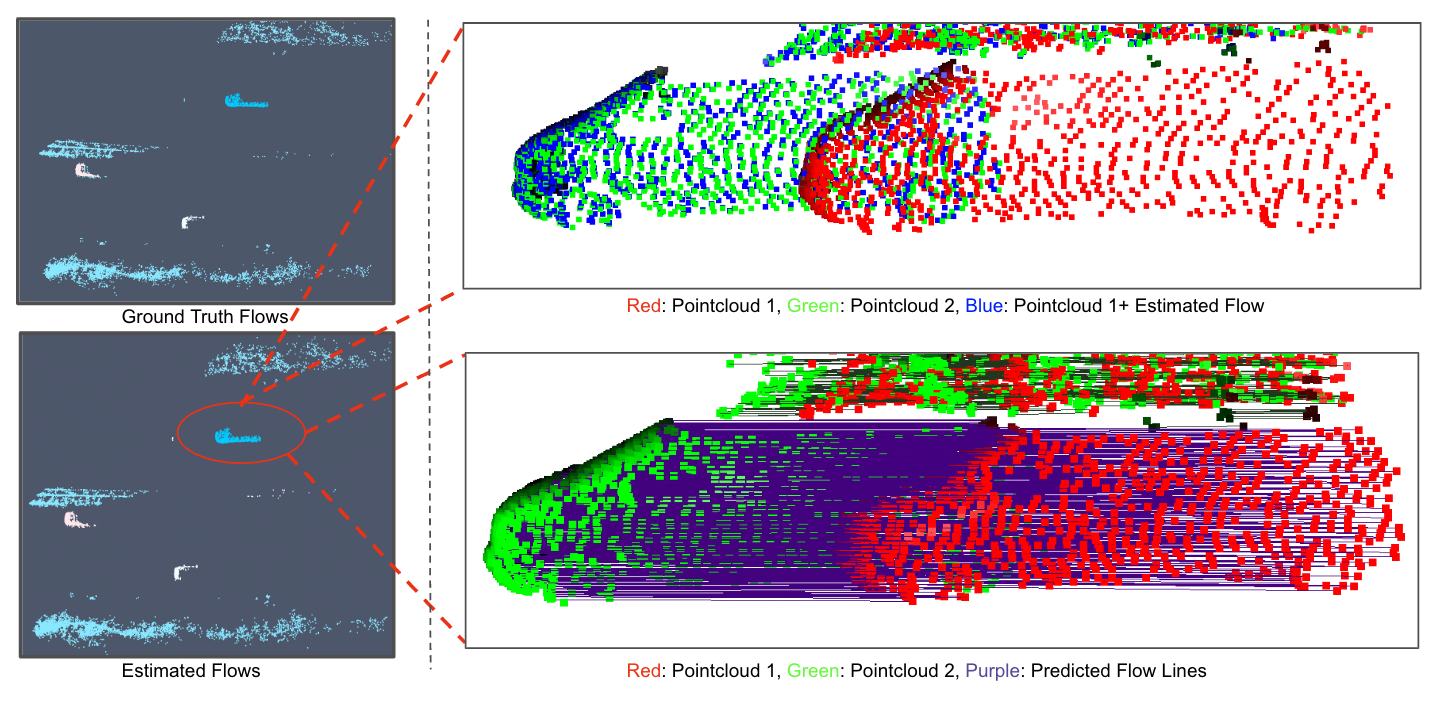}
\vspace{-10pt}
\caption{\textbf{Visualization of predicted flows on the KITTI Dataset. } \textbf{Left:} The color-coded map illustrates a comparison between ground truth flow vectors and our predicted flow values. \textbf{Top Right}: The point cloud \textcolor{red}{$P_{T-1}$} is depicted in red, point cloud \textcolor{green}{$P_{T}$} in green, and the translated point cloud \textcolor{blue}{$(P_{T-1}+F)$} in blue. Note the proximity of the blue points to the ground truth green points, indicating high prediction accuracy. \textbf{Bottom Right}: The visualization of predicted flow lines is presented.}
\label{fig:flow_vis_1}
\end{figure*}

\section{Related Work}
\paragraph{Non-learning-based methods}
A method revolving around the analysis of a sequence of stereo images~\cite{vedula1999three} first pioneered the field of scene flow estimation. They first calculated the 2D optical flow between each pair of stereo images, which gives the motion of pixels in the image plane. They then used the epipolar geometry between the stereo pairs to back-project the 2D optical flow to the 3D scene flow.
Another non-learning-based iterative method is the Non-rigid Iterative Closest Point (NICP) work by Amberg \etal \cite{amberg}. The goal of this method was to register the point cloud representation of the scene to the current frame. However, this method was sensitive to initialization and not suitable for large-scale differences between the template and the scanned mesh. 

\paragraph{Supervised learning-based methods}
The most prevalent algorithms in learning-based approaches for scene flow estimation involve training a flow regressor model, typically a neural network, to compute flow vectors between point clouds in the ambient 3D space \cite{flownet3d, flownet3d++, camliflow, whatmatters, pointnet, pointnet++, festa}. The introduction of deep learning architectures for point cloud processing, such as PointNet \cite{pointnet} and its hierarchical extension PointNet++ \cite{pointnet++} by Qi \etal, has inspired numerous works on scene flow estimation. Learning based state-of-the-art models have proposed innovative network architectures that build on these foundational models to estimate scene flow directly from point clouds, such as FlowNet3D \cite{flownet3d}.

FlowNet3D \cite{flownet3d} employed PointNet++ for feature encoding of points and introduced a flow embedding layer that learned to aggregate geometric similarities and spatial relations of points for motion estimation. Wang \etal \cite{flownet3d++} subsequently enhanced FlowNet3D by incorporating geometric constraints in the form of point-to-plane distance and angular alignment, resulting in FlowNet3D++. Numerous models have since emerged, focusing on improving point cloud feature extraction and adding a flow refinement module on top of it.

One such model is FESTA \cite{festa}, which introduced spatial and temporal features through an attention mechanism that effectively captures the temporal relationships between consecutive point clouds and the spatial relationships. BiPointFlowNet \cite{cheng2022bi}, another notable model, leverages both forward and backward correspondence information in a bidirectional approach to better handle occlusions and out-of-view regions. The FLOT3D \cite{flot3d} method introduced a novel frustum-based optimization method for scene flow estimation, leveraging learned optimization techniques and applying a series of transformations in a coarse-to-fine manner. Lastly, FH-Net \cite{fhnet} presents a fast hierarchical network that addresses the computational challenges associated with dense 3D scene flow estimation, utilizing a hierarchical structure and lightweight building blocks to exploit both local and global features.

\paragraph{Self-supervised learning-based methods}

Recent research has seen a surge of interest in self-supervised learning as a potential solution for enhancing scene flow estimation from point cloud data \cite{pointpwc, Mittal_2020_CVPR, Slim, scoop, rigidflow, superpoints, selfpointflow} and monocular images \cite{selfsupervised, mono1, mono2}. PointPwcNet \cite{pointpwc} introduced the concept of cycle consistency loss, which in turn inspired Mittal et al. \cite{Mittal_2020_CVPR} to apply a similar approach for point cloud correspondence identification. They innovatively merged cycle consistency loss with nearest neighbor loss to tackle the challenge of scene flow estimation. PointPwcNet \cite{pointpwc} made use of the Chamfer Distance \cite{chamfer}, smoothness constraints, and Laplacian regularization to train scene flow in a self-supervised way.
SLIM\cite{Slim} solved scene flow estimation while simultaneously classifying motion segmentation. Flowstep3d \cite{kittenplon2021flowstep3d}, on the other hand, employed a soft point matching module to calculate pairwise matches between points in the source and target point clouds. Based on this, we introduced a local correlation weight matrix in our algorithm to improve soft correspondences within our objective function. These strategies have demonstrated considerable promise in augmenting scene flow estimation and can be customized to various datasets while preserving generalizability. Nevertheless, self-supervised learning-based methods still demand considerable training data to reach satisfactory learning outcomes, and the associated training cost can often be exorbitantly high.

\paragraph{Optimization based methods}

Optimization-based methods present a distinct class of non-learning-based approaches for estimating scene flow. Uniquely, these methods circumvent the need for model training, opting instead for complete runtime flow optimization. Such an approach was notably employed by authors in \cite{graphprior}, where they encoded the prior of the flow to be as rigid as possible by minimizing the graph laplacian defined over the source points. A subsequent work by Argo AI \cite{nsfp} replaced the explicit graph with a neural prior using a coordinate-based MLP, thereby implicitly regularizing the optimized flow field.  Recently, the scene flow estimation method, SCOOP \cite{scoop}, was introduced, which innovatively combines pre-training on a subset of data to learn soft correspondences and secure initial flows, followed by optimization-based flow refinement steps. This hybrid approach has allowed SCOOP to deliver competitive results, using considerably less training data. 

\section{Method}

\begin{figure*}[t!]
\centering
\includegraphics[width=\linewidth]{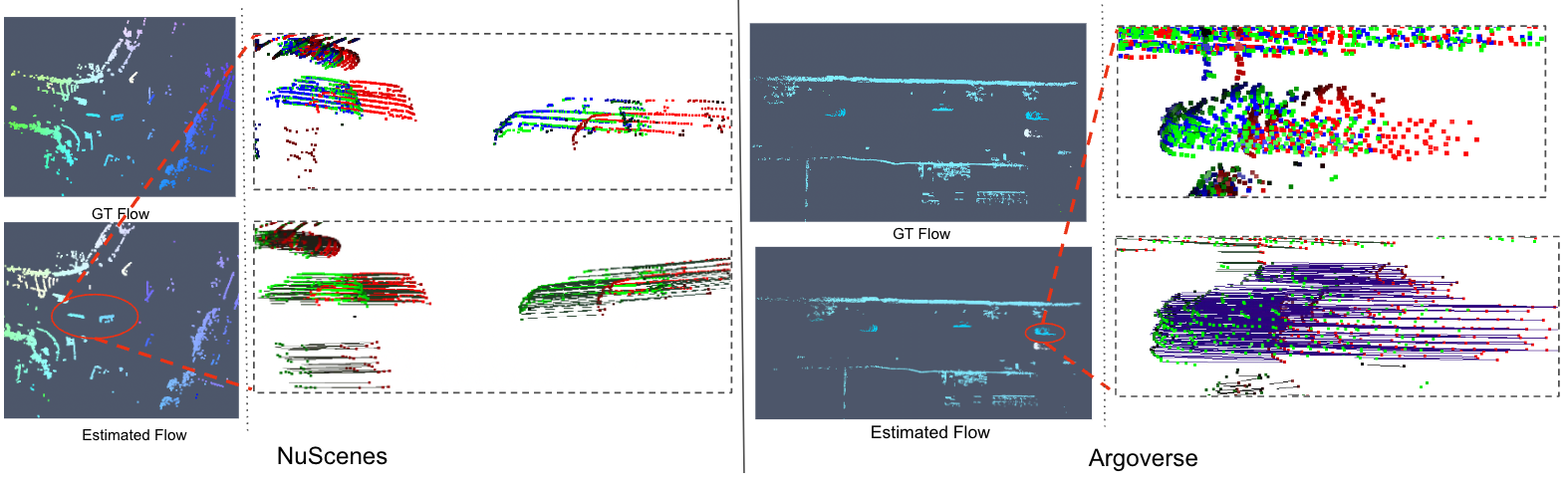}
\vspace{-10pt}
\caption{\textbf{Visualization of predicted flows on the nuScenes and Argoverse datasets.} \textbf{Left:} A color-coded map illustrates a comparison between the ground truth flow vectors and our predicted flow values. \textbf{Top Right}: The point cloud \textcolor{red}{$P_{T-1}$} is depicted in red, point cloud \textcolor{green}{$P_{T}$} in green, and the transformed point cloud \textcolor{blue}{$P_{T-1}+F$} in blue. \textbf{Bottom Right}: The visualization of predicted flow lines is presented. The nuScenes example is particularly intricate, as the ego-vehicle is positioned at a turn, resulting in angled flows. Additionally, the color-coded map illustrates the variance in flow values associated with each point.}
\label{fig:flow_vis_2}
\end{figure*}

\textbf{Problem definition}: Given two sets of 3D point clouds, $P_{t-1} \in \mathbb{R}^{n_1 \times 3}$  and $P_{t} \in \mathbb{R}^{n_2 \times 3}$, representing a dynamic scene at two different times $t-1$ and $t$, respectively, the task is to compute the 3D motion vector for each point in $P_{t-1}$.

Since the number of points in each set may be different and there may not be a one-to-one correspondence between points, we model the motion of each point in $P_{t-1}$ using a flow vector $f$ in $R^3$, and the collection of these points is called a flow field $F \in \mathbb{R}^{n_1 \times 3}$.

In our approach, we predict flow $f_i$ for each point $p_i$ in $P_{t-1}$ such that the total distance between 2 sets of point clouds is minimum. Thus, we want to find an optimal set of flow vectors such that:

\begin{equation}
    F^* = \argmin_{F}\sum_{p_i \in P_{t-1}} dist(p_i+ f_i, P_t)
\end{equation}
where $dist$ is the function which computes distance between $p_i+f_i$ and its corresponding nearest point in $P_t$. 

\subsection{Integrated Ego-motion Compensation} \label{pointcloudmatch}

In real-world scenarios, LiDAR data is often captured from a moving vehicle, making ego-motion capture critical for localization and motion estimation. Moreover, with the majority of points in autonomous driving datasets being static, the incorporation of ego motion can simplify scene flow estimation by aligning the input point cloud pairs and focusing on dynamic objects.
Though standard datasets such as Argoverse \cite{argoverse} and nuScenes \cite{nuscenes} provide ego-motion information, this data hasn't been tapped in previous scene flow estimation research. Furthermore, errors can occur due to the absence or inaccuracy of ego-motion sensor data in certain scenarios.
To diminish dependence on external odometry data, we integrate an Iterative Closest Point (ICP) \cite{ICP} based transformation function, denoted as T, into our optimization process, enabling its estimation alongside flow vector estimation. At each step, we transform the current point cloud into the next coordinate frame, minimising the distance between corresponding static points. 

The updated objective function is defined as:



\begin{equation}
F^* = \argmin_{F, T}\sum_{p_i \in P_{t-1}} dist(Tp_i+ f_i, P_t).
\end{equation}

In addition to point cloud flow $F$,  we also estimate the transformation $T$ from the frame of point cloud 1 to the frame of point cloud 2. This is to delineate the transformation of static parts of the scene due to vehicle motion from the flow of dynamic objects and their associated points.
In section \ref{ablation}, we demonstrate the effectiveness of this transformation in enhancing accuracy.

\subsection{Local Correlation Weight Matrix} \label{correlation}

To establish the flow field, $F$, between two point clouds, it is essential to find correspondences between points in the source and target point clouds. Previous methods, such as \cite{Mittal_2020_CVPR, graphprior}, use the 1 nearest neighbor approach to find correspondences, which can lead to sub-optimal solutions due to initial inaccuracies.

Building on the global correlation unit in FlowStep3D \cite{kittenplon2021flowstep3d}, which employs cosine similarity to find soft correlations between features, we introduce a local correlation weight matrix. For each point, $p_i$, in the source point cloud, $P_{t-1}$, transformed as $(Tp_i+f_i)$, we identify the $k_{local}$ nearest points in the target point cloud, $P_t$. We then compute a similarity score, $sim_{ij}$, based on the exponential of the negative squared distance, $d_{ij}$, between point pairs:

\input{iccv2023AuthorKit/tables/kitti}

\begin{equation}
sim_{ij}=e^{-d^2_{ij}}.
\end{equation}

The correlation weight, $M_{ij}$, representing the confidence in the similarity between the source and target points, is calculated using $sim_{ij}$:

\begin{equation}
W_{ij}= \exp\left(\frac{sim_{ij}-1}{\epsilon}\right).
\end{equation}

Using these weights, we compute a weighted average of the target points $q_j \in P_t$ to determine the optimal correspondence, $q_{{avg}_i}$, for each source point, $p_i$:

\begin{equation}
q_{{avg}_i}= \frac{\sum_{j=1}^{k_{local}}W_{i,j}q_j}{\sum_{j=1}^{k_{local}}W_{i,j}}.
\end{equation}

This leads to a new objective function:

\begin{equation}
E_{fit}= \sum_{i=1}^{n_1} ||p_{i} + f_i - q_{i}||^2_2.
\label{eq:egomotion}
\end{equation}

We use our objective function in a bidirectional manner to both sets of point clouds, to effectively align it with the principles of Chamfer Distance \cite{chamfer}, ensuring a more symmetrical and comprehensive evaluation of correspondences.

\textbf{Note:} $\epsilon$ and $K_{local}$ are hyperparameters. While $\epsilon$ is fixed at 0.03, $K_{local}$ depends on the density of the dataset, with larger values for sparse datasets and smaller values for dense datasets.

\subsection{Adaptive Distance Threshold} 

While constructing the local correlation weight matrix by finding the $K$ nearest neighbors, it is critical to eliminate outliers that could adversely affect point correspondences. Although some studies employ a fixed distance limit of 2.0m, we introduce an adaptive distance threshold that evolves during optimization. 

This adaptive threshold, $d_{thresh}$, decreases at regular intervals of $n$ steps, with the rationale that the flow vectors’ accuracy improves with each iteration, necessitating a more restrictive threshold for better matching precision. The threshold is halved at each interval until it reaches a lower limit of 0.2m. Correlations between points exceeding this distance threshold are assigned a weight of zero in the weight matrix $M$, ensuring that only those within the threshold contribute to the final flow field.

In the current implementation, the interval is empirically set to 100 steps, which means that the threshold is halved after every 100 iterations.

\subsection{Rigidity Constraint}

Inspired by \cite{graphprior} and building upon the principles presented in it, we add a rigidity constraint in our optimization objective to maintain geometric coherence in the source point cloud. Specifically, we enforce local rigidity among points in close proximity within a subgraph of $K_{rigid}$ points, mimicking the characteristics of rigid body motion.

Our rigidity constraint is enforced by minimizing the difference between the flows of each pair of points within the subgraph. The formulation leverages the graph laplacian, which implicitly captures the topological structure of the point cloud, to regularize the scene flow. Mathematically, the rigidity function \(E_{rigid}\) is expressed as:

\begin{equation}
E_{rigid}(f_i) = \sum_{i,j \in S} W_{rigid}^{ij} \left\| f_i - f_j \right\|^2_2,
\end{equation}
where \(S\) is the set of edges of a subgraph \(G\) containing \(K_{rigid}\) points in the source point cloud \(P_{T-1}\), and \(W_{rigid}^{ij}\) is a weight defined similar to $sim_{ij}$:

\begin{equation}
W_{rigid}^{ij} = e^{-d_{ij}^2},
\end{equation}
where \(d_{ij}\) is the distance between points \(p_i\) and \(p_j\) in subgraph \(G\). The weight \(W_{rigid}^{ij}\) assigns higher importance to pairs of points that are closer, promoting rigidity within the local region. 

\textbf{Note:} We use $K_{rigid}=50$, as proposed in \cite{graphprior}.
\\~\\
Combining all the proposed methods above, our final objective function becomes:

\begin{align}
\label{eq:objective}
    E_{obj} & = E_{fit} + \alpha_{rigid}E_{rigid} \\
     &= \sum_{i=1}^{n_1} ||Tp_{i} + f_i - q_{avg}||^2_2 +  \alpha_{rigid}  \sum_{i,j \in S} W_{rigid}^{ij} \left\| f_i - f_j \right\|^2_2
\end{align}
where $n_1$ represents all points in point cloud $P_{t-1}$ and $\alpha_{rigid}$ is the weight of the rigidity loss.
\input{iccv2023AuthorKit/tables/results2048}
\section{Experiments}

In this section, we evaluate OptFlow's performance on synthetic and real-life autonomous driving datasets and compare it with recent state-of-the-art methods for scene flow estimation. Additionally, we compare their generalizability, speed of execution, and model complexity. Our less complex method achieves competitive performance without any annotations and training data available.

\paragraph{Datasets:}
These are the four major datasets that we evaluated our models on:

\begin{enumerate}
    \item \textbf{FlyingThings3D}: FlyingThings3D is a large-scale synthetic dataset of random objects from the ShapeNet collection. Similar to \cite{graphprior, nsfp}, we use a pre-processed dataset released by \cite{flownet3d}. The dataset is evaluated on 2000 test samples. 
    \item \textbf{KITTI Scene Flow:}
    KITTI was designed to evaluate scene flow methods on real-world self-driving scenarios. In our experiments, we use a pre-processed dataset that was released by \cite{graphprior}. The dataset is split into 100 train and 50 test sample sets, with ground points filtered out.
    \item \textbf{nuScenes:}
    nuScenes dataset is a large-scale autonomous driving dataset in urban environments. It is challenging due to the presence of a lot of dynamic objects in the scene and also because of the presence of occlusions in the LiDAR point clouds. As there are no official scene flow annotations, we use the ego-vehicle poses and 3D object tracks to create pseudo-labels as done in \cite{graphprior, nsfp}. The ground points below a certain threshold are also filtered out. The results are evaluated on 310 test samples from 150 test scenes, same as \cite{graphprior, nsfp}.
    \item \textbf{Argoverse:} Argoverse is another large-scale challenging autonomous driving dataset released by Argo AI. Similar to nuScenes, scene flow annotations are not provided for Argoverse and the same process has been followed to derive the pseudo labels. The results are evaluated on 212 test samples, same as \cite{graphprior, nsfp}.
\end{enumerate}

\paragraph{Metrics:} To assess the effectiveness of our method, we utilized commonly used metrics (such as those in previous works \cite{flownet3d, nsfp, graphprior, Mittal_2020_CVPR, pointpwc}) which include: EPE $\varepsilon$ (end-point error), which measures the mean absolute distance between two point clouds; $Acc_5$ (Accuracy Strict), which calculates the percentage of estimated flows where the EPE  is less than 0.05m or the relative error $E_0$ is less than 5\%; $Acc_{10}$ (Accuracy Relax), which measures the percentage of estimated flows where the EPE is less than 0.1m or the relative error $E_0$ is less than 10\%; and $\theta$, the mean angle error between the estimated and ground-truth scene flows.

\paragraph{Implementation Details:}
Our implementation leverages PyTorch, utilizing its automatic differentiation library to optimize our objective function with the AdamW optimizer. We initialize our flow parameters, $F \in \mathbb{R}^{n_1 \times 3}$, as empty tensors. Additionally, we initialize a rotation vector, $r \in \mathbb{R}^{1 \times 3}$, and a translation vector, $t \in \mathbb{R}^{1 \times 3}$—components of the transformation matrix $T \in \SE{3}$  defined in \eqref{eq:egomotion}—using values derived from ICP registration. The learning rate is set at 4e-3, and the optimization process is run for 600 iterations, incorporating early stopping based on loss value. To compare with leading architectures, we sample our dataset using 2048 points, 8192 points, and the full point cloud. All experiments are executed on an NVIDIA T4 GPU.


\section{Results}

\subsection{Comparison with different methods}
 As demonstrated in Table \ref{tbl:kitti}, we assess our method using both 2048 points and the entire point cloud limited to a depth of 35m, aligning our evaluation with parameters typically reported in the current literature. The results depicted in Table \ref{tbl:kitti} are either obtained from the recently published study \cite{scoop} or independently reproduced by our team.

In Table \ref{tab:main_table}, our experimental setup follows the methodologies adopted by \cite{graphprior, nsfp}. Here, we consider the full point cloud without imposing the range constraints applied in the previous evaluation. These evaluation metrics are sourced from \cite{nsfp}.

A thorough examination of both tables reveals that our method consistently outperforms alternative techniques across all primary datasets. Specifically focusing on non-learning-based approaches, OptFlow surpasses the performance of all comparable methods across every dataset, while also delivering the fastest inference time, as demonstrated in Figure \ref{fig:speed_vs_acc}.

Figures \ref{fig:flow_vis_1} and \ref{fig:flow_vis_2} provide qualitative results of our algorithm applied to the KITTI, Argoverse, and nuScenes datasets.

\subsection{Inference Time analysis and tradeoff}
\begin{figure}[t!]
\centering
\includegraphics[width=\linewidth]{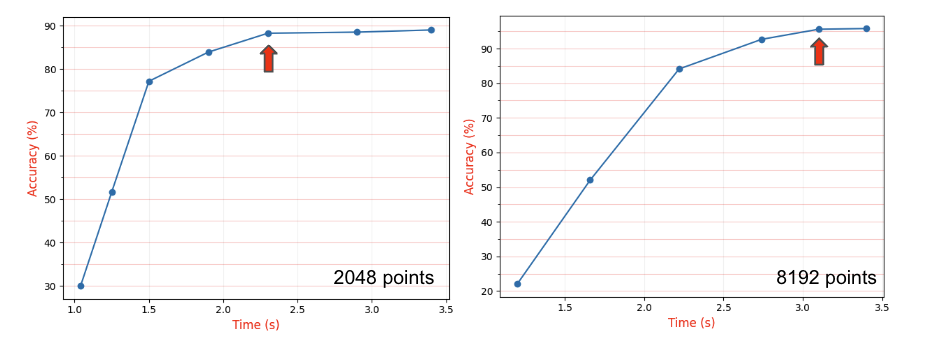}
\vspace{-10pt}
\caption{\textbf{Performance of OptFlow algorithm on KITTI with different iterations and time taken for them. The red arrow shows the most optimal performance.}}
\label{fig:time_vs_}
\end{figure}

Figure \ref{fig:time_vs_} presents a detailed analysis of our OptFlow model's performance over a range of timesteps on the KITTI dataset. Optimum performance is observed at the 600th iteration, with inference times of approximately 2.3 seconds for 2048 points and 3.1 seconds for 8192 points.

\begin{figure}[t!]
\centering
\includegraphics[width=\linewidth]{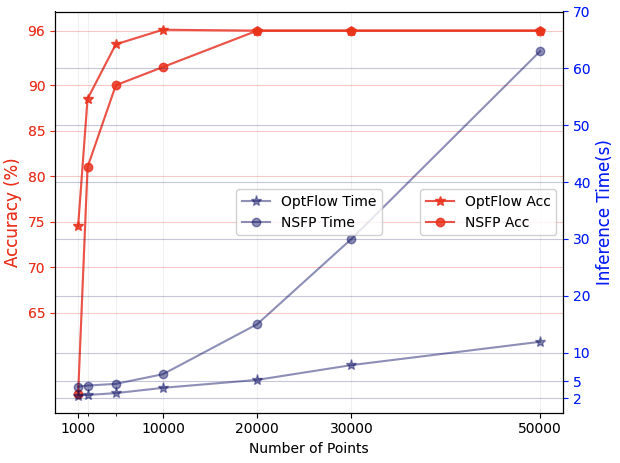}
\vspace{-10pt}
\caption{\textbf{Performance($Acc_{5}$) comparison of our algorithm and NSFP on the KITTI dataset for varying point cloud densities. This shows we get around 7x speedup over NSFP \cite{nsfp} as the point cloud density increases.} The point clouds are processed parallelly after 8k points as discussed in sec. 1 of supplementary.}
\label{fig:teaser_scoop}
\end{figure}


\subsection{Evaluation on high-density point clouds}

As depicted in Tables~\ref{tbl:kitti} and \ref{tab:main_table}, we predominantly evaluated our model's performance using 2048 points. Nonetheless, it's important to note that real-world LiDAR sensor data often contain tens of thousands of points, necessitating the testing of scene flow methods on such high-density point clouds.

Our model's performance on high-density point clouds is evaluated in Figure \ref{fig:teaser_scoop}, which illustrates the trade-off between accuracy and inference time as we scale the number of points within the KITTI Scene Flow dataset. For benchmarking, we've compared our method with Neural Scene Flow Prior (NSFP) \cite{nsfp}, presently the best-performing algorithm for non-learning-based approaches.

The results clearly demonstrate that our method substantially outperforms Neural Scene Flow Prior \cite{nsfp} in terms of accuracy, even while operating at a faster inference speed and utilizing fewer points.

\subsection{Ablation Studies}
\label{ablation}

\begin{table}[h]
\small
\centering
\begin{tabular}{@{ }  p{0.7\linewidth} c c @{ }}
\toprule
Experiment & $EPE\!\downarrow$ & $\%_{5}\!\uparrow$ \\
\midrule
(a) W/O integrated ego-motion transformation & 0.081 & 91.6 \\
(b) W/O adaptive distance threshold \\ (thresh = 2m)      & 0.035 & 91.9 \\
(c) W/O correlation weight matrix   & 0.029 & 95.6 \\
\textbf{Our complete method  (a + b + c)}                                   & \textbf{0.028} & \textbf{95.8} \\
\bottomrule
\end{tabular}
\caption{\textbf{Ablation experiment with 8192 points} evaluated on KITTI\textsubscript{t}.}
\label{tbl:ablation_8192}
\end{table}

Tables \ref{tbl:ablation_8192} and \ref{tbl:ablation_2048} demonstrate the influence of each component presented in our paper. Table \ref{tbl:ablation_8192} quantifies the percentage change each element induces in the complete algorithm. For evaluation, we used a certain setup: considering our full method as a baseline and illustrating the impact of each constituent on the $EPE$ and $Acc_{5}$ metrics. We carried out experiments which entailed a) the omission of integrated ego-motion transformation, thus forgoing the transformation matrix and assuming the identity matrix for $T$ in equation \ref{eq:egomotion}; b) keeping a fixed distance threshold of 2m, with no adaptive adjustments; c) disregarding the correlation weight matrix, hence only the nearest neighbor in the target point-cloud was contemplated as a corresponding point for each source point; d) implementing our comprehensive algorithm with all components included.

Table \ref{tbl:ablation_8192} highlights the fact that every component makes a significant contribution, though the correlation weight matrix has a minor variance from the full method. In contrast, Table \ref{tbl:ablation_2048} shows that when we decrease the number of points, the impact of each component amplifies considerably, especially the correlation weight matrix, which increases $Acc_{5}$ by a significant 4\%.

\begin{table}[t!]
\small
\centering
\begin{tabular}{@{} p{0.70\linewidth} cc @{}}
\toprule
Experiment & $EPE\!\downarrow$ & $\%_{5}\!\uparrow$ \\
\midrule
(a) W/O integrated ego-motion transformation & 0.095 & 84.12 \\
(b) W/O adaptive distance threshold \\(thresh = 2m) & 0.063 & 79.56 \\
(c) W/O correlation weight matrix & 0.050 & 84.90 \\
\textbf{Our complete method  (a + b + c)} & \textbf{0.049} & \textbf{88.25} \\
\bottomrule
\end{tabular}
\caption{\textbf{Ablation experiment with 2048 points} evaluated on KITTI\textsubscript{t}.}
\label{tbl:ablation_2048}
\end{table}

\section{Applications}

\begin{figure}[t!]
\centering
\includegraphics[width=\linewidth]{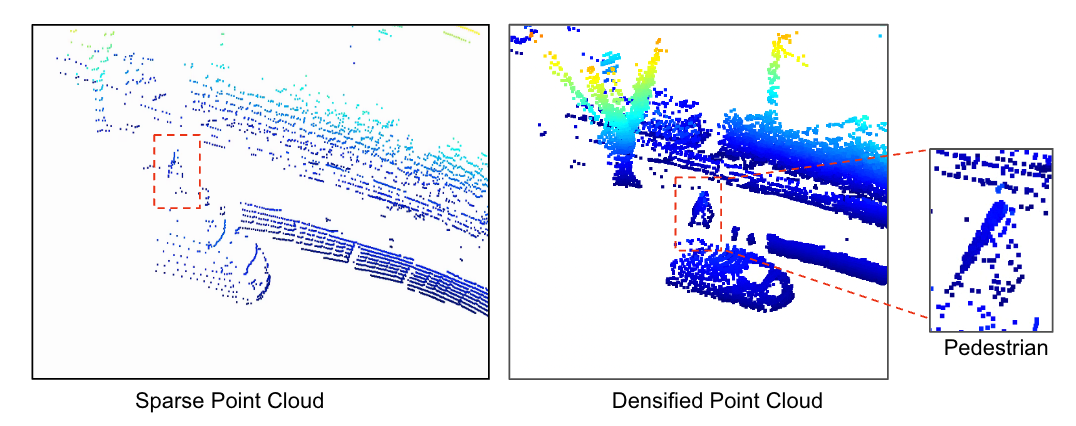}
\vspace{-10pt}
\caption{\textbf{Point Cloud Densification:} One example from nuScenes depicting the application of point cloud densification. Our method was able to densify a pedestrian in a sparse point cloud.}
\label{fig:densify}
\end{figure}

\textbf{Densification:} Our scene-flow estimation methods demonstrate such precision that they can be employed to densify successive point-cloud frames. By computing pairwise scene-flow estimates, we can project the current point-clouds - augmented with estimated flows - onto the succeeding point-cloud, and perpetuate this densification process through `$n$' frames. As illustrated in Figure \ref{fig:densify}, we apply our method on the nuScenes dataset to successfully densify a pedestrian situated within an extremely sparse point cloud. 

\begin{figure}[t!]
\centering
\includegraphics[width=\linewidth]{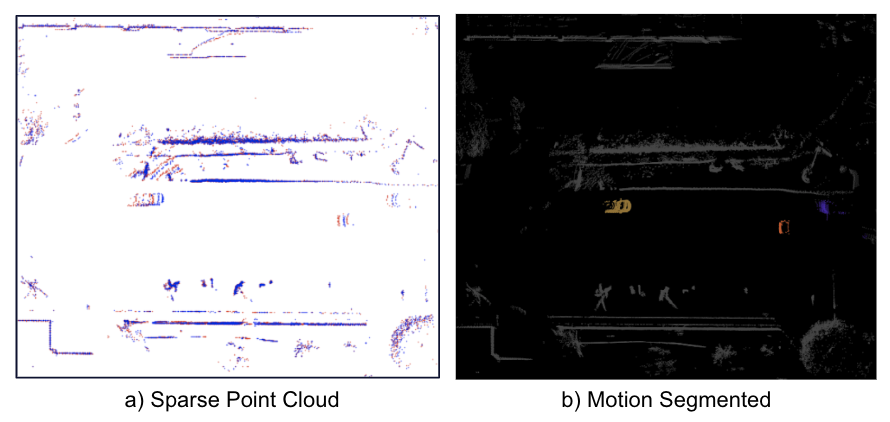}
\vspace{-10pt}
\caption{\textbf{Motion Segmentation(BeV):} On the left, a pair of sparse point clouds is shown in red and blue, after applying ego-motion compensation. On the right, motion is segmented out and dynamic objects are color-coded, just from the scene flow result. Points exceeding a set threshold are then marked as dynamic.}
\label{fig:motion_segment}
\end{figure}

\textbf{Object Annotation/Motion Segmentation:}
The incorporation of ego-motion compensation into our objective function facilitates accurate alignment between two input pairs of point clouds, assisting in closely aligning static objects. This alignment process diminishes the flow values of static objects, thus simplifying the differentiation between dynamic and static objects. As shown in Figure \ref{fig:motion_segment}, motion segmentation is achieved after the ego-motion compensation transformation is applied to an input pair of point clouds. Furthermore, our algorithm lends itself to the annotation of dynamic objects in new datasets, extending its utility beyond scene flow estimation.


\section{Limitations}

As reflected in Table \ref{tbl:kitti} and Figure \ref{fig:speed_vs_acc}, our method achieves state-of-the-art results with better inference speed amongst all non-learning-based techniques. However, the execution time may not suffice for real-time applications, such as for use in autonomous vehicles.

Another challenge lies in the need for precision in hyperparameter settings for each specific dataset. When handling a new dataset, our method requires a hyperparameter tuning algorithm to obtain peak performance. Hyperparameters, such as $\alpha_{rigidity}$ in the final objective function \eqref{eq:objective} and $K_{local}$ in the local correlation weight matrix, are pivotal in influencing the performance across each dataset.

\section{Conclusion}
To conclude, our research introduces an innovative and efficient non-learning scene flow estimation method. Our method incorporates a local correlation weight matrix and an adaptive distance threshold, both instrumental in accelerating flow field convergence and enhancing correspondence accuracy.

Additionally, we proposed an intrinsic ego-motion compensation function that bolsters precision and curbs computational complexity by reducing flow value computations for static points and primarily concentrating on dynamic points. This approach culminates in state-of-the-art results in the Accuracy Strict ($Acc_5$) metric for all key autonomous driving datasets, and the fastest inference time among all non-learning methods.

Finally, we demonstrate the versatility of our algorithm with its applications in dynamic/static object annotation and point cloud densification.

{\small
\bibliographystyle{ieee_fullname}
\bibliography{egbib}
}

\newpage
\end{document}


\title{OptFlow: Fast Optimization-based Scene Flow Estimation without Supervision \\ Supplementary Material }

\author{Rahul Ahuja\quad Chris Baker\quad Wilko Schwarting\\
ISEE AI\\
{\tt\small \{rahulahuja, chrisbaker, wilko\}@isee.ai}}
\maketitle

\input{iccv2023AuthorKit/tables/results2048}

\section{Faster OptFlow in Batches}

\begin{figure}[t!]
\centering
\includegraphics[width=\linewidth]{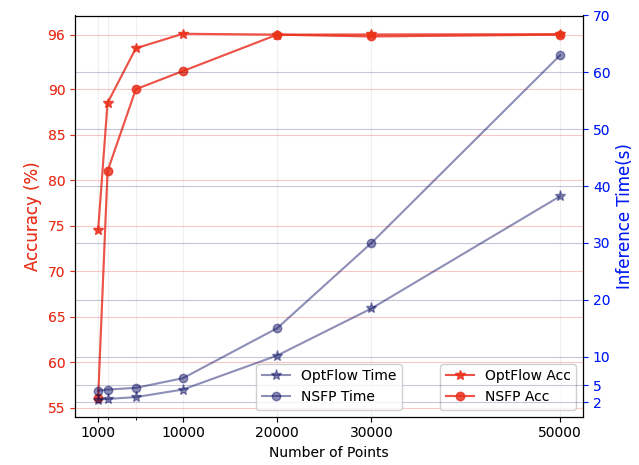}
\vspace{-10pt}
\caption{\textbf{Performance comparison of our algorithm and NSFP on the KITTI dataset for varying point cloud densities computed as a whole point cloud together.}}
\label{fig:points_vs_time_1}
\end{figure}
\begin{figure}[t!]
\centering
\includegraphics[width=\linewidth]{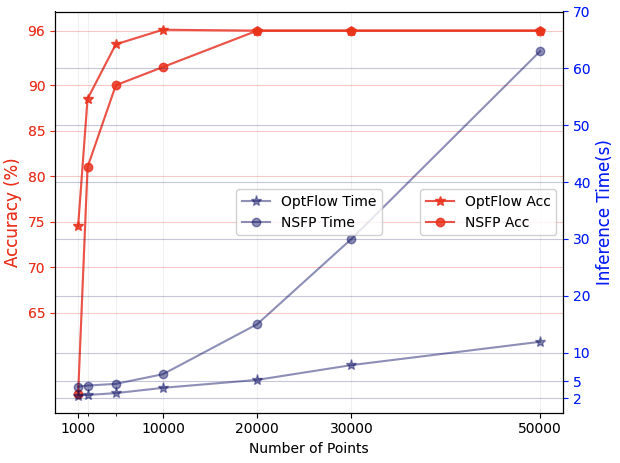}
\vspace{-10pt}
\caption{\textbf{Performance comparison of our algorithm and NSFP on the KITTI dataset for varying point cloud densities computed in parallel batches of 8192 points. This shows we get around 7x speedup over NSFP \cite{nsfp} as the point cloud density increases.}}
\label{fig:points_vs_time_2}
\end{figure}

The KNN search between the source and target point clouds within the fit function loss constitutes a significant bottleneck, increasing the inference time of our model. As the number of points grows, so does the search space, thereby leading to an extended time for the KNN operation to find the best correspondences. Importantly, as depicted in Figure \ref{fig:points_vs_time_1} from above, our model's performance regarding $Acc_{5}$ tends to level off around 8k-10k points. As a result of this observation, we chose to process our point clouds in parallel batches of 8192 points. This approach involves randomly sampling sets of 8192 points, batching them together, and running our scene flow estimation algorithm on them in parallel. Upon completion, we collate the results for final evaluation. By limiting the search space for each KNN operation, this method substantially cuts down on processing time while preserving the accuracy of our scene flow estimates. The revised correlation between accuracy and time, adjusted for different point cloud densities, is presented in Figure \ref{fig:points_vs_time_2}. As indicated by the figure, this strategy allows our model to operate seven times faster than NSFP \cite{nsfp}, significantly enhancing its efficiency.

\begin{figure*}[t!]
\centering
\includegraphics[width=\linewidth]{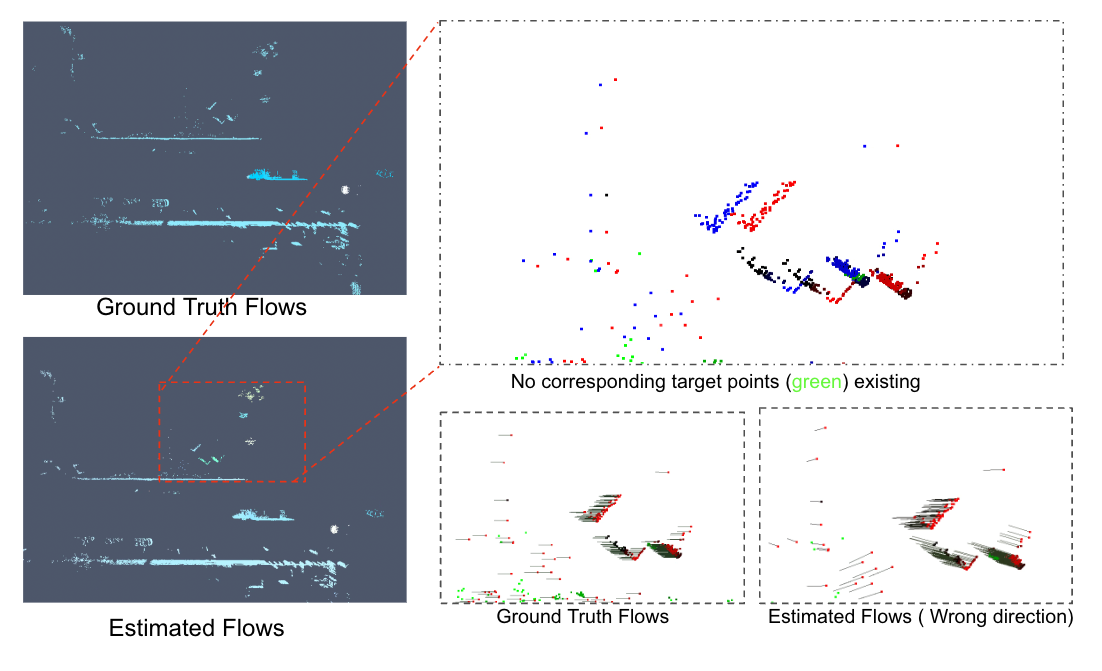}
\vspace{-10pt}
\caption{\textbf{Failure case due to missing points:} The source point cloud is represented in red, the target point cloud in green, and the warped point cloud in blue. We can observe that the flow estimates for the static points are incorrect. This is due to the lack of corresponding points, which prevents our model from forming accurate scene flow estimations for these points. Despite this, the results are not significantly compromised thanks to our \textbf{Adaptive Distance Threshold} feature. This component reduces the limit for finding correspondences, thereby diminishing the likelihood of numerous false positives.}
\label{fig:failure}
\end{figure*}

\section{Real-Time OptFlow}

The original implementation of OptFlow typically takes around 1.5 to 2 seconds to converge. However, its suitability for real-time applications depends on the specific use case, accuracy requirements, and End Point Error (EPE) constraints. To explore its real-time potential, we conducted an experiment using sequential data from the NuScenes dataset \cite{nuscenes}, with a time interval of 0.2 seconds between frames.

In this experiment, we initially processed the first two consecutive point clouds, running our optimization algorithm for a full 500 iterations. We then initialized the flow values for the next pair of point clouds using nearest neighbor matches from the previously estimated flows. In dynamic scenes, scene flow vectors for objects in consecutive frames tend to be similar or exhibit slight changes if objects are in motion. This allows us to reduce the number of iterations required for subsequent runs while maintaining good accuracy.

For the consecutive runs, we executed our algorithm for just 30-40 iterations, taking approximately 0.17-0.2 seconds to complete. Importantly, we retained the flow vectors and the ICP transformation function from the previous run. Although we lacked ground truth flow annotations for these samples, we devised an evaluation metric called End Point Error Difference (EPED), calculated as follows: $EPED = EPE(flow_{500}, flow_{30})$. Here, $flow_{500}$ represents a run with 500 iterations, while $flow_{30}$ represents flow values obtained from 30 iterations.

Remarkably, our experiments yielded an $EPED$ value of only 0.053 meters. This implies that by reducing the number of iterations and achieving an inference speed of 200 milliseconds for OptFlow, we only compromised 0.05 meters of EPE accuracy, making it a viable option for real-time applications.

This real-time OptFlow optimization will be further explored in our future work.

\section{Failure Cases}

Our method, which relies on the nearest neighbor search in the fit function, can potentially encounter failure cases. Specifically, this can occur when no correspondence exists between a point in the source point cloud and another point in the target point cloud. Under such circumstances, the loss function may falsely identify a match, leading to false positives. This scenario is frequently encountered with static background points such as trees, poles, and occluded objects as these often present missing points in subsequent point clouds. The problem is amplified when we evaluate using point clouds of 2048 or 8192 points, given that points are randomly sampled. When the point cloud is sparse, this may lead to the generation of false positives. Nevertheless, as point density increases, the risk of this failure decreases, enhancing our evaluation results.

\begin{figure*}[t!]
\centering
\includegraphics[width=\linewidth]{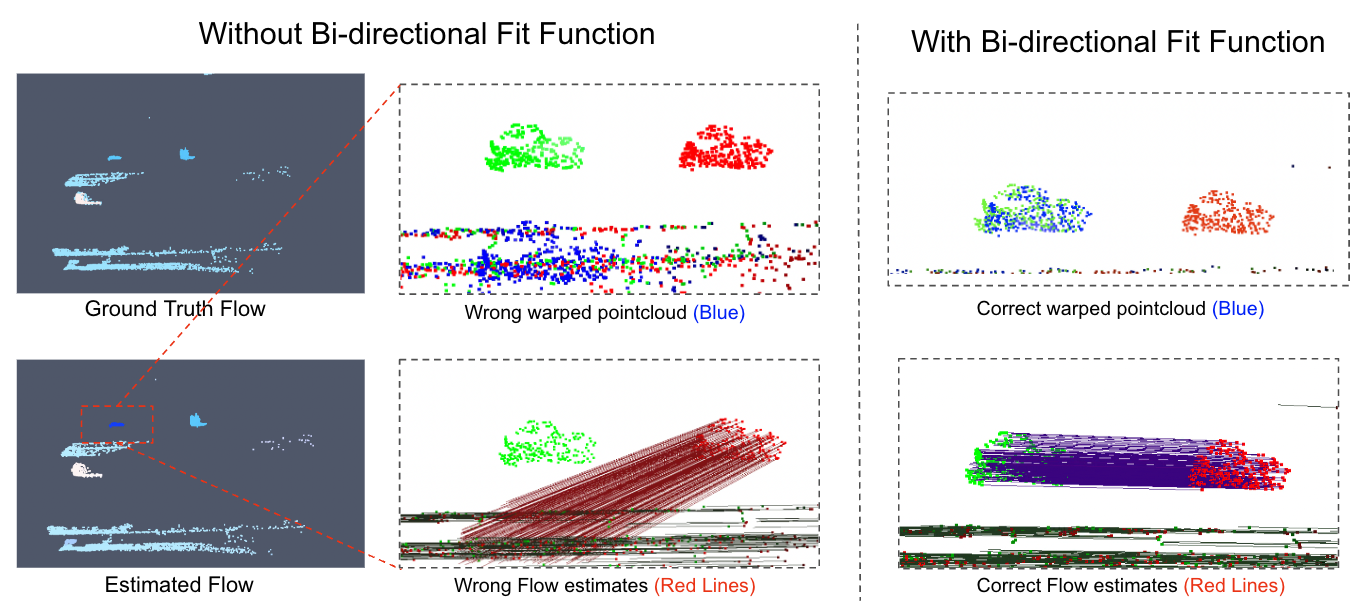}
\vspace{-10pt}
\caption{\textbf{Impact of the bi-directional Fit Function:} The source point cloud is depicted in red, the target point cloud in green, and the warped point cloud in blue. On the \textbf{left}, we observe incorrect scene flow estimates that do not align with the ground truth flow values. In the \textbf{center} image, the assigned correspondences are inaccurate, resulting in a misplaced warped point cloud. On the \textbf{right}, we see the dramatic improvement achieved by applying the bi-directional fit function. The correspondences are significantly corrected and closely match the target point cloud.}
\label{fig:bidirect}
\end{figure*}

Figure \ref{fig:failure} exemplifies the impact of these missing points using a scene from the Argoverse dataset. In the color-coded map, our scene flow estimations for the area marked in red are observed to deviate from the ground truth flows. Upon closer examination of these points, we find that corresponding points are absent in the target point cloud, as shown by the absence of green points. The estimated flows for these points are visually represented.

Despite the flaws in our flow angles, the flow estimates are not excessively erroneous. This is attributable to the adaptive distance threshold. In the limitations highlighted by works such as \cite{graphprior} and \cite{scoop}, these warped point clouds ($P_{T-1} + F$) would have shrunk towards the nearest available point. However, in our case, the adaptive distance threshold curtails the impact of false positives, hence yielding better estimates.

\subsection{FlyingThings3D}

\begin{figure}[t!]
\centering
\includegraphics[width=\linewidth]{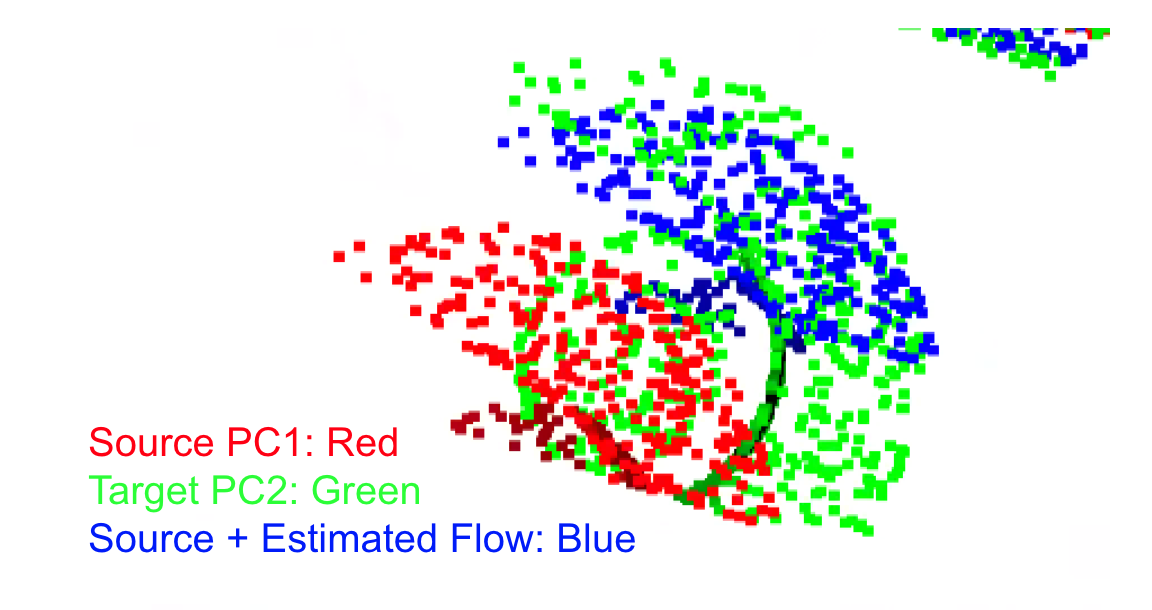}
\vspace{-10pt}
\caption{\textbf{FlyingThings3D failure} This is a part of an object from a scene in the FlyingThings3D dataset. The \textcolor{red}{red} points denote the source point cloud, while the \textcolor{green}{green} points represent the target point cloud. Additionally, the \textcolor{blue}{blue} points illustrate the source point cloud after integrating estimated flows. A notable observation is that the \textcolor{red}{source} point cloud contains fewer points, indicating partial visibility compared to the \textcolor{green}{target} point cloud. Consequently, our algorithm estimates scene flow for the reduced set of points in the source cloud, leading to missing flow estimations for numerous points in the target point cloud. This disparity in flow estimations results in higher End Point Error (EPE) values and lower accuracy scores in such scenarios. }
\label{fig:flying3d}
\end{figure}

Continuing with the discussion of challenging scenarios, the FlyingThings3D dataset presents an interesting case due to its frequent occlusions and partial visibility. It's worth noting that our algorithm exhibits lower performance in terms of End Point Error compared to learning-based methods that were specifically trained on the FlyingThings3D dataset. As previously explained, our reliance on nearest neighbor loss means that in cases of occlusions and limited visibility, the assigned nearest correspondence may not be correct. An example provided below illustrates a situation where the source point cloud, when combined with the estimated flow, misses the target point cloud.

Change in visibility is a common challenge faced by optimization-based methods. However, despite these difficulties, our approach still manages to achieve a remarkable 12\% improvement in $Acc_{5}$ compared to the current state-of-the-art optimization-based method \cite{nsfp}. This demonstrates the significant contributions our methodology brings to the field, even in demanding scenarios.

\section{Bi-directional fit function}

As outlined in our primary submission, we apply our fit function \ref{eq:egomotion} in a bi-directional manner. 
\begin{equation}
E_{fit}= \sum_{i=1}^{n_1} ||Tp_{i} + f_i - q_{{avg}_i}||^2_2.
\label{eq:egomotion}
\end{equation}

This approach mimics the effect of the Chamfer Distance \cite{chamfer}. This original metric is a nearest neighbor loss function, which identifies one nearest neighbor for each point in the source point cloud $P_{T-1}$ in relation to the target point cloud $P_{T}$, and subsequently calculates the mean squared difference between the two. This process is then reciprocated, with the source serving as the target and vice versa. The mean of these two error terms constitutes the Chamfer loss.

In line with this, our approach initially designates $P_{T-1}$ as the source and $P_{T}$ as the target. In the second round, these roles are reversed with $P_{T}$ serving as the source and $P_{T-1}$ as the target point cloud. The corresponding $q_{avg}$ is determined through the locally correlated weight matrix applied to each target point cloud.

Table \ref{tbl:bidirect} presents the performance of our model when using the bi-directional fit function compared to a unidirectional application. As the results indicate, the bidirectional application of our fit function yields notably improved performance.

In figure \ref{fig:bidirect}, we also depict how the flow estimates fail without the bi-directional fit function.

\begin{table}[h]
\small
\centering
\begin{tabular}{ @{ }lcccc@{ } }
\toprule
Experiment & $EPE\!\downarrow$ & $\%_{5}\!\uparrow$ &$\%_{10}\!\uparrow$ & $\theta_{\epsilon}$\\
\midrule
(a) Bi-directional fit loss & 0.049 & 88.25 & 95.1 & 0.13 \\
(b) Non Bi-directional fit loss   & 0.073 & 85.26 & 92.7 & 0.16 \\
\bottomrule
\end{tabular}
\caption{\textbf{Effect of Bi-directional fit function on KITTI using 2048 points.}}
\label{tbl:bidirect}
\end{table}

\section{Implementation Details}

In this section, we share the hyperparameters that resulted in the outcomes presented in our paper. We employed RayTune \cite{raytune} and Hyperopt \cite{Hyperopt} for an extensive hyperparameter search. Throughout our experiment, we maintained $K_{rigid}$ at a constant value of 50, as this yielded the most optimal results and was recommended in \cite{graphprior}.

The parameters $\alpha_{rigid}$ and $K_{local}$, which are used in the nearest neighbor search for the local target correlation matrix, played a vital role in achieving the desired outcomes. Table \ref{tbl:Parameters_Setting} outlines the hyperparameters applied in our experiment.

We advise conducting your own hyperparameter search to identify the most suitable solution for your specific scenario.

\begin{table}[h]
\small
\centering
\begin{tabular}{ @{ }lcc@{ } }
\toprule
Dataset & $K_{local}$ & $\alpha_{rigid}$\\
\midrule
(a) FlyingThings3D\cite{ft3d} & 12 & 14.2  \\
(b) NuScenes \cite{nuscenes} & 20 & 19.6  \\
(c) KITTI \cite{kitti}& 15 & 9.57 \\
(d) Argoverse \cite{argoverse}& 15 & 19.2\\
\bottomrule
\end{tabular}
\caption{\textbf{Hyperparameters used for each dataset.}}
\label{tbl:Parameters_Setting}
\end{table}

\begin{figure}[t!]
\centering
\includegraphics[width=\linewidth]{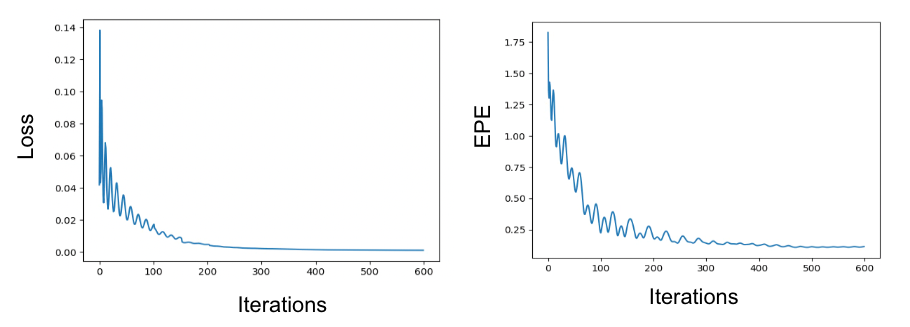}
\vspace{-10pt}
\caption{\textbf{Left: }Loss curve vs iterations, \textbf{Right: }EPE vs iterations}
\label{fig:loss}
\end{figure}

\section{Visualizations}

\subsection{Comparison of OptFlow with Baseline}

\begin{figure}[t!]
\centering
\includegraphics[width=\linewidth]{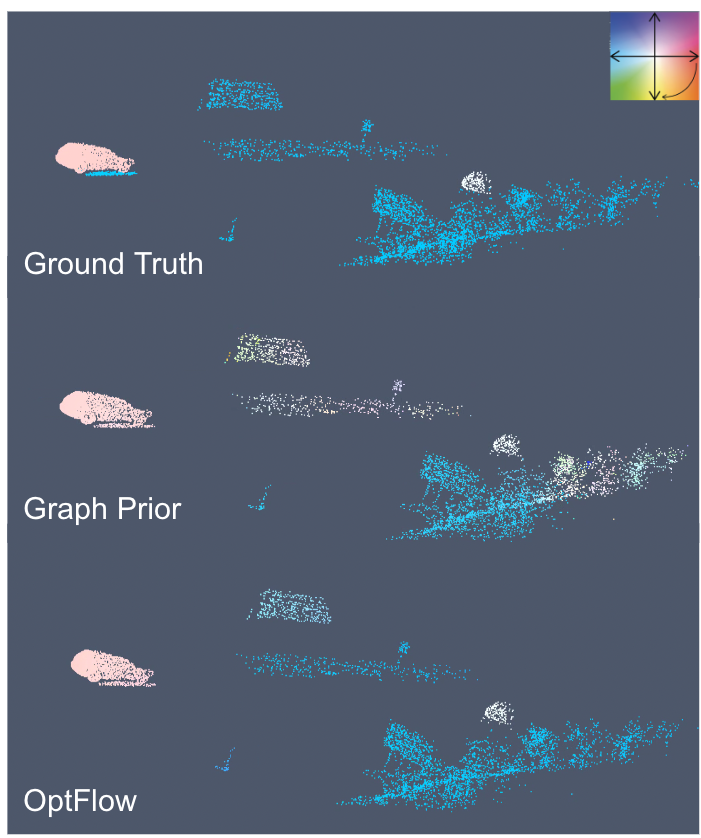}
\vspace{-10pt}
\caption{\textbf{Scene Flow Visualization of Graph Prior \cite{graphprior} Method vs. OptFlow (Ours)}: Our method's estimated scene flow, particularly in the background, presents a more accurate representation compared to the Graph Prior method. The ground truth scene flow aligns more closely with our estimation.}
\label{fig:baseline}
\end{figure}

In Figure \ref{fig:baseline}, we contrast our proposed OptFlow method with the baseline method, Graph Prior~\cite{graphprior}. It becomes evident that our flow estimates significantly outperform those produced by Graph Prior, particularly in correctly predicting flow values for static background points. This comparison clearly demonstrates the advantages of OptFlow in terms of accuracy in scene flow prediction.

\subsection{Effect of integrated ego-motion compensation}

In many state-of-the-art scene flow estimation algorithms, the primary focus is typically on dynamic objects, with static objects or points often receiving less attention. This approach may work satisfactorily for dense datasets such as KITTI, but it can result in a high likelihood of false positives when dealing with sparse datasets.

The motivation for our research to incorporate an integrated ego-motion compensation was to mitigate these false positives. By aligning the static points effectively, our algorithm is better positioned to accurately identify the scene flow estimates of dynamic objects.

\begin{figure}[t!]
\centering
\includegraphics[width=\linewidth]{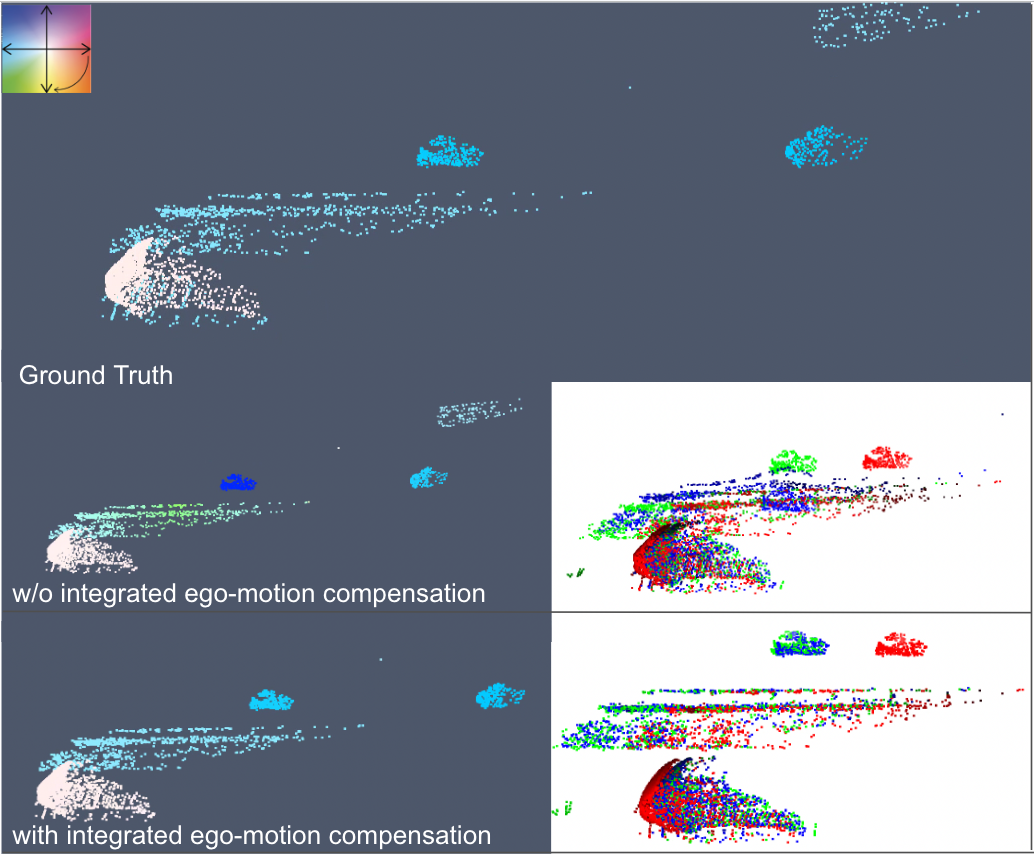}
\vspace{-10pt}
\caption{\textbf{Effect of integrated ego-motion compensation within our objective function}: The transformation function aids in more accurate point cloud alignment, significantly reducing the likelihood of false positives, as demonstrated above.}
\label{fig:ego-fig}
\end{figure}

As depicted in Figure \ref{fig:ego-fig}, the inclusion of an integrated ego-motion compensation transformation function significantly reduces false positives and yields estimates that are more closely aligned with the ground truth flow values. 

\subsection{Tough Example: Ego vehicle surrounded by dynamic objects}

\begin{figure*}[t!]
\centering
\includegraphics[width=\linewidth]{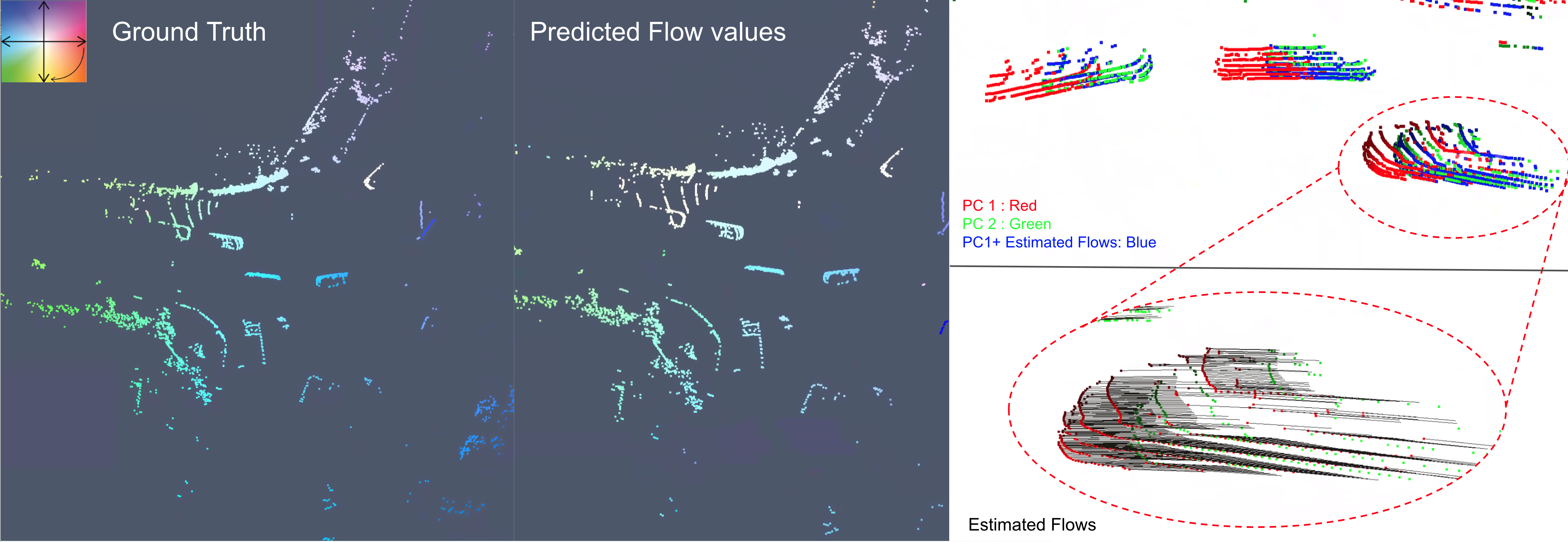}
\vspace{-10pt}
\caption{\textbf{Predicting Scene Flow in a Dynamic Scene}: This visualization depicts the estimated scene flow surrounding an ego vehicle. The left image uses color gradients to indicate flow velocity and direction. Similar color values depict similarity in flow values. The image on the right depicts estimated flow vectors (\textcolor{blue}{blue}) onto the actual target point cloud data (\textcolor{green}{green}). The proximity of the predicted and target points demonstrates the accuracy of the scene flow model in this busy urban setting.}
\label{fig:toughexample}
\end{figure*}

In figure \ref{fig:toughexample}, we provide an enlarged view of Figure 3 from the main paper. This visualization illustrates a challenging scenario where the ego vehicle is navigating a turn and is surrounded by numerous dynamic objects. Remarkably, our algorithm continues to deliver excellent performance even in these complex conditions.

\newpage

{\small
\bibliographystyle{ieee_fullname}
\bibliography{egbib}
}

%% file: rvc-notation.tex
\usepackage{amsmath}
\usepackage{amssymb}
\usepackage{accents}
\usepackage{bm}
\usepackage{xifthen}
\usepackage{color}
\usepackage{fancyvrb}
\usepackage{graphicx,scalerel}

\newcommand{\ba}{\begin{eqnarray}}
\newcommand{\ea}{\end{eqnarray}}

\newcommand{\presup}[1]{\,{}^{\scriptscriptstyle #1}\!}

\newcommand{\pose}[1][_NONE]{\ifthenelse{\equal{#1}{_NONE}}{}{\presup{#1}}{\mathbf{\xi}}}
\newcommand{\estpose}[1][_NONE]{\ifthenelse{\equal{#1}{_NONE}}{}{\presup{#1}}{\hat{\mathbf{\xi}}}}
\newcommand{\hpose}[1][_NONE]{\ifthenelse{\equal{#1}{_NONE}}{}{\presup{#1}}{\hat{\mathbf{\xi}}}}
\newcommand{\posedot}[1][_NONE]{\ifthenelse{\equal{#1}{_NONE}}{}{\presup{#1}}{\mathbf{\nu}}}

\newcommand{\q}[1][_NONE]{\ifthenelse{\equal{#1}{_NONE}}{}{\presup{#1}}{\mathring{q}}}
\newcommand{\uquat}[2][_NONE]{\ifthenelse{\equal{#1}{_NONE}}{}{\presup{#1}}{\mathring{#2}}}
\newcommand{\duquat}[2][_NONE]{\ifthenelse{\equal{#1}{_NONE}}{}{\presup{#1}}{\dot{\mathring{#2}}}}
\newcommand{\quat}[2][_NONE]{\ifthenelse{\equal{#1}{_NONE}}{}{\presup{#1}}{\breve{#2}}}

\DeclareMathAlphabet{\mathitbf}{OML}{cmm}{b}{it}
\newcommand{\twist}[2][_NONE]{\ifthenelse{\equal{#1}{_NONE}}{}{\presup{#1}}{\mathcal{S}}}
\renewcommand{\vec}[2][_NONE]{\ifthenelse{\equal{#1}{_NONE}}{}{\presup{#1}}{\boldsymbol #2}}

\newcommand{\hvec}[2][_NONE]{\ifthenelse{\equal{#1}{_NONE}}{}{\presup{#1}}{\tilde{\vec{#2}}}}
\newcommand{\evec}[2][_NONE]{\ifthenelse{\equal{#1}{_NONE}}{}{\presup{#1}}{\hat{\vec{#2}}}}
\newcommand{\bvec}[2][_NONE]{\ifthenelse{\equal{#1}{_NONE}}{}{\presup{#1}}{\bar{\vec{#2}}}}

\newcommand{\dhvec}[2][_NONE]{\ifthenelse{\equal{#1}{_NONE}}{}{\presup{#1}}{\dot{\tilde{\vec{#2}}}}}
\newcommand{\dvec}[2][_NONE]{\ifthenelse{\equal{#1}{_NONE}}{}{\presup{#1}}{\dot{\vec{#2}}}}
\newcommand{\ddvec}[2][_NONE]{\ifthenelse{\equal{#1}{_NONE}}{}{\presup{#1}}{\ddot{\vec{#2}}}}

\newcommand{\mat}[2][_NONE]{\ifthenelse{\equal{#1}{_NONE}}{}{\presup{#1}\,}{{\mathbf #2}}}
\newcommand{\dmat}[2][_NONE]{\ifthenelse{\equal{#1}{_NONE}}{}{\presup{#1}\,}{{\dot{\mathbf #2}}}}
\newcommand{\emat}[2][_NONE]{\ifthenelse{\equal{#1}{_NONE}}{}{\presup{#1}\,}{\hat{\mathbf#2}}}
\newcommand{\matfn}[3][_NONE]{\ifthenelse{\equal{#1}{_NONE}}{}{\presup{#1}}{{\mat{#2}}\left(#3\right)}}
\newcommand{\Rt}[2][_NONE]{\ifthenelse{\equal{#1}{_NONE}}{}{\presup{#1}}{{\bf R}\left(#2\right)}}

\newcommand{\point}[2][_NONE]{\ifthenelse{\equal{#1}{_NONE}}{}{\ensuremath{\presup{#1}}}{\ensuremath{\mathrm{#2}}}}

\newfont{\School}{pncr}
\newfont{\eightTR}{pncr at 8pt}

\fvset{formatcom={\color{blue}\baselineskip=-\maxdimen\lineskip=0pt},xleftmargin=4mm,commentchar=!}

\DefineVerbatimEnvironment{Code}{Verbatim}{}

\DefineVerbatimEnvironment{CodeSmall}{Verbatim}{fontsize=\scriptsize,xleftmargin=1mm,commentchar=!}

\DefineVerbatimEnvironment{CodeNum}{Verbatim}{numbers=left,numbersep=4pt,fontsize=\footnotesize,xleftmargin=4mm}

\newcommand{\func}[2][_NONE]{\ifthenelse{\equal{#1}{_NONE}}{\index{code}{#2}}{\index{code}{#1}}\ifthenelse{\boolean{draft}}{{\color{green}\Verb+#2+}}{\Verb+#2+}}

\newcommand{\methodb}[2]{\index{code}{#1@\textbf{#1}!.#2}\ifthenelse{\boolean{draft}}{{\color{magenta}\Verb+#1.#2+}}{\Verb+#1.#2+}}
\newcommand{\method}[2]{\index{code}{#1@\textbf{#1}!.#2}\ifthenelse{\boolean{draft}}{{\color{magenta}\Verb+#2+}}{\Verb+#2+}}
\newcommand{\class}[1]{\index{code}{#1@\textbf{#1}}\ifthenelse{\boolean{draft}}{{\color{cyan}\Verb+#1+}}{\Verb+#1+}}
\newcommand{\property}[1]{\index{property}{#1}\ifthenelse{\boolean{draft}}{{\color{cyan}\Verb+#1+}}{\Verb+#1+}}

\newcommand{\SE}[1]{\ensuremath{\mathrm{{\bf SE}(#1)}}}

\usepackage{xparse}

\NewDocumentCommand{\ovec}{ o m o }{%
  \IfValueT{#1}{\presup{#1}}%
  \vec{#2}%
  \IfValueT{#3}{_{#3}}
  ^{\scalebox{0.6}{\#}}%
  }
\NewDocumentCommand{\omat}{ o m o }{%
  \IfValueT{#1}{\presup{#1}}%
  \mat{#2}%
  \IfValueT{#3}{_{#3}}
  ^{\scalebox{0.6}{\#}}%
  }

\NewDocumentCommand{\obspose}{ o o }{%
  \IfValueT{#1}{\presup{#1}}%
  \mathbf{\xi}%
  \IfValueT{#2}{_{#2}}
  ^{\scalebox{0.6}{\#}}%
  }

%% file: iccv2023AuthorKit/tables/kitti.tex
\begin{table*}[t!]
\small
\centering
\begin{tabular}{@{ } l l l c @{ } c @{ } c @{ } c @{ } c @{ }}
\toprule
Method & Supervision & Learning Based &\#points &  $\:EPE\!\downarrow$ & $\%_5\uparrow$ & $\%_{10}\uparrow$ & $\:Out.\!\downarrow$ \\
\midrule
Just Go With The Flow (JGF)~\cite{Mittal_2020_CVPR} & \textit{Full} + \textit{Self} + \textit{Self} & True & 2048 & 0.105 & 46.5 & 79.4 & - \\
Self Point-Flow (SPF)~\cite{SPF} & \textit{Self} + \textit{Self} &True & 2048  & 0.089 & 41.7 & 75.0 & - \\
RigidFlow~\cite{rigidflow} & \textit{Self} & True & 2048  & 0.117 & 38.8 & 69.7 & - \\
SCOOP~\cite{scoop} & \textit{Self} & Partial & 2048  &
{0.052} &  {80.6} &  {92.9} &  {19.7} \\
Neural Prior~\cite{nsfp} & \textit{Self} & False & 2048 & {0.052} &  84.8 & 94.3 & \textbf{16.4} \\
\textbf{OptFlow (Ours)} & \textit{Self} & False & 2048  & \textbf{0.049} & \textbf{88.25} & \textbf{95.1} & 18.2 \\
\midrule
Graph Prior~\cite{graphprior} & \textit{Self} & False & Full Pointcloud  & 0.082 & 84.0 & 88.5 & - \\
Neural Prior~\cite{nsfp} & \textit{Self} & False & Full Pointcloud  &  {0.034} & 92.3 & 96.4 & \textbf{12.0} \\
SCOOP\textsuperscript{+}~\cite{scoop} & \textit{Self} & Partial & Full Pointcloud  &
0.039 &  {93.6} &  {96.5} &  {15.2}\\
\textbf{OptFlow (Ours)} & \textit{Self} &False & Full Pointcloud  & \textbf{0.028} & \textbf{96.1} & \textbf{97.8} & 13.2 \\
\bottomrule
\end{tabular}
\caption{\textbf{Quantitative comparison on 2048 points and full pointcloud on KITTI Dataset with 35m range set on the point cloud.} Comparison with prior work as reported in the \cite{scoop}. Here $EPE$ denotes the end-point error, $\%_{5}$ denotes strict accuracy,$\%_{10}$ denotes relaxed accuracy and $Out.$ denotes the percentage of points that are outliers (i.e.$ EPE\geq$0.3m)}
\label{tbl:kitti}
\end{table*}

%% file: iccv2023AuthorKit/tables/results2048.tex
\newcommand{\tabitem}{~~\llap{\textbullet}~~}
\begin{table*}[t!]
\begin{adjustbox}{width=\textwidth}
\begin{tabular}{@{}lrcccccccccccccccccc@{}}
\toprule
 &\multicolumn{4}{c}{\thead{\textbf{FlyingThings3D}~\cite{ft3d} \\ \small \textit{\#Test: 2,000}}} &\hphantom &\multicolumn{4}{c}{\thead{\textbf{KITTI Scene Flow}~\cite{kitti} \\ \small \textit{\#Test: 50}}} &\hphantom &\multicolumn{4}{c}{\thead{\textbf{Argoverse Scene Flow}~\cite{argoverse} \\ \small \textit{\#Test: 212}}} &\hphantom &\multicolumn{4}{c}{\thead{\textbf{nuScenes Scene Flow}~\cite{nuscenes} \\ \small \textit{\#Test: 310}}} \\
\cmidrule{2-20} 
&\thead{$EPE$ \\ $(m)$} &\thead{$\%_5\uparrow$ \\ $(\%)$} &\thead{$\%_{10}\uparrow$ \\ $(\%)$} &\thead{$\theta_{\epsilon}\downarrow$ \\ $(rad)$} &&\thead{$EPE$ \\ $(m)$} &\thead{$\%_5\uparrow$ \\ $(\%)$} &\thead{$\%_{10}\uparrow$ \\ $(\%)$} &\thead{$\theta_{\epsilon}\downarrow$ \\ $(rad)$} &&\thead{$EPE$ \\ $(m)$} &\thead{$\%_5\uparrow$ \\ $(\%)$} &\thead{$\%_{10}\uparrow$ \\ $(\%)$} &\thead{$\theta_{\epsilon}\downarrow$ \\ $(rad)$} &&\thead{$EPE$ \\ $(m)$} &\thead{$\%_5\uparrow$ \\ $(\%)$} &\thead{$\%_{10}\uparrow$ \\ $(\%)$} &\thead{$\theta_{\epsilon}\downarrow$ \\ $(rad)$} \\ \toprule
\tabitem FlowNet3D~\cite{flownet3d} &0.134 &22.64 &54.17 &0.305     && 0.199 &10.44 &38.89 &0.386     &&0.455 &1.34 &6.12 &0.736       &&0.505 &2.12 &10.81 &0.620 \\
\tabitem PointPWC-Net~\cite{pointpwc} &\textbf{0.121} &{29.09} &\textbf{61.70} &\textbf{0.229}      && {0.142} & {29.91} & {59.83} & {0.239}    && {0.405} & {8.25} & {25.47} & {0.674}    && {0.442} & {7.64} & {22.32} & {0.497} \\
\arrayrulecolor{black}\bottomrule
\tabitem JGF~\cite{Mittal_2020_CVPR} &\multicolumn{4}{c}{---}   &&0.218 &10.17 &34.38 &0.254     &&0.542 &8.80 &20.28 &0.715        &&0.625 &6.09 &0.139 &0.432 \\
\tabitem PointPWC-Net~\cite{pointpwc} &\multicolumn{4}{c}{---}     &&0.177 &13.29 &42.15 &0.272    &&0.409 &9.79 &29.31 &0.643    &&0.431 &6.87 &22.42 &0.406 \\
\arrayrulecolor{black}\bottomrule
\tabitem Non-rigid ICP~\cite{amberg} &0.339 &14.05 &35.68 &0.480     &&0.338 &22.06 &43.03 &0.460    &&0.461 &4.27 &13.90 &0.741     &&0.402 &6.99 &21.01 &0.492 \\
\tabitem{Graph Prior}~\cite{graphprior} &{0.259} &{16.30} & {41.60} & {0.369}     && {0.093} & {64.76} & {82.13} & {0.137}   && {0.257} & {25.26} & {47.50} & {0.467}    && {0.288} & {20.19} & {43.59} & {0.337} \\
\tabitem{{NSFP}}~\cite{nsfp} &0.234 &19.16 &46.74 &0.341 && 0.050 &81.68 &93.19 &0.133 && \textbf{0.159} &38.43&63.08 &\textbf{0.374} && \textbf{0.175}  &35.18 &63.45&0.279 \\
\tabitem{\textbf{OptFlow (ours)}} & {0.224} & \textbf{31.73} & {57.1}  & {0.340} && {0.052} & \textbf{84.3} & \textbf{93.2} &\textbf{0.130} && {0.20}  &\textbf{43.85} &\textbf{65.5}&{0.39} &&{0.216}&\textbf{40.85}&\textbf{65.97} &\textbf{0.271}\\
\arrayrulecolor{black}\bottomrule
\end{tabular}
\end{adjustbox}
\vspace{0.01ex}
\caption{\textbf{Results on 2048 points with no point cloud range limit set.} Comparison with prior work as reported in the NSFP \cite{nsfp}. Top section between shows \textit{off-the-shelf} \textbf{supervised learning methods}; the middle section shows \textbf{self-supervised learning methods}; and the bottom section shows \textbf{non-learning-based methods}. Here $EPE$ denotes the end-point error, $\%_{5}$ denotes strict accuracy,$\%_{10}$ denotes relaxed accuracy and $\theta_{\epsilon}$ denotes angle error.}
\label{tab:main_table}
\end{table*}